\documentclass[10pt,twocolumn,letterpaper]{article}

\usepackage{cvpr}              % To produce the CAMERA-READY version
\usepackage{times}
\usepackage{epsfig}
\usepackage{float}
\usepackage{multirow}
\usepackage[accsupp]{axessibility}

\usepackage{graphicx}
\usepackage{amsmath}
\usepackage{amssymb}
\usepackage[pagebackref=true,breaklinks=true,letterpaper=true,colorlinks,bookmarks=false]{hyperref}

\usepackage[capitalize]{cleveref}
\crefname{section}{Sec.}{Secs.}
\Crefname{section}{Section}{Sections}
\Crefname{table}{Table}{Tables}
\crefname{table}{Tab.}{Tabs.}
\renewcommand\footnotemark{}

\def\iccvPaperID{4872} % *** Enter the ICCV Paper ID here

\begin{document}

%%%%%%%%% TITLE - PLEASE UPDATE
\title{U-RED: Unsupervised 3D Shape Retrieval and Deformation \\ for Partial Point Clouds}

\author{Yan Di$^{1*}$, Chenyangguang Zhang$^{2*}$, Ruida Zhang$^{2*}$, Fabian Manhardt$^{3}$, 
Yongzhi Su$^{4}$,\\
Jason Rambach$^{4}$, Didier Stricker$^{4}$,
Xiangyang Ji$^{2}$ and Federico Tombari$^{1,3}$\\
\textsuperscript{1}Technical University of Munich, \textsuperscript{2}Tsinghua University,
\textsuperscript{3} Google,
\textsuperscript{4} DFKI\\
\tt\small{\{yan.di@, tombari@in.\}tum.de},
\tt\small{\{zcyg22, zhangrd21\}@mails.tsinghua.edu.cn}
\\
\thanks{*Authors with equal contributions.}
\thanks{Codes: \url{https://github.com/ZhangCYG/U-RED}}
}
\maketitle

%%%%%%%%% ABSTRACT
\begin{abstract}
%3D perception of object shapes from sensor data is an essential task in semantic scene understanding.
%However, due to (self-)occlusion and inherent sensor limitations, the collected data is often incomplete.
%Moreover, the lack of labelled real-world data limits the application potential.
In this paper, we propose \textbf{U-RED}, an \textbf{U}nsupervised shape \textbf{RE}trieval and \textbf{D}eformation pipeline that takes an arbitrary object observation as input, typically captured by RGB images or scans, and jointly retrieves and deforms the geometrically similar CAD models from a pre-established database to tightly match the target.
%We train U-RED on the synthetic datasets and directly apply it to the real-world scenario.
% To handle partial point clouds, we develop an unsupervised collaborative learning strategy and exploit the geometry consistency between partial shape and full shape.
Considering existing methods typically fail to handle noisy partial observations, U-RED is designed to address this issue from two aspects.
First, since one partial shape may correspond to multiple potential full shapes, the retrieval method must allow such an ambiguous one-to-many relationship.
Thereby U-RED learns to project all possible full shapes of a partial target onto the surface of a unit sphere.
Then during inference, each sampling on the sphere will yield a feasible retrieval.
Second, since real-world partial observations usually contain noticeable noise, a reliable learned metric that measures the similarity between shapes is necessary for stable retrieval.
In U-RED, we design a novel point-wise residual-guided metric that allows noise-robust comparison.
%A collaborative learning scheme emphasizing geometric consistencies is further proposed to improve robustness against noise.
Extensive experiments on the synthetic datasets PartNet, ComplementMe and the real-world dataset Scan2CAD demonstrate that U-RED surpasses existing state-of-the-art approaches by 47.3$\%$, 16.7$\%$ and 31.6$\%$ respectively under Chamfer Distance.
% Codes and trained models will be released soon.

% When applied to the real-world dataset Scan2CAD, Un-3RD achieves a remarkable 30.5$\%$ leap forward over competitors.
\end{abstract}

%%%%%%%%% BODY TEXT
\section{Introduction}
\begin{figure}[t]
    \centering
    \includegraphics[width=0.43\textwidth]{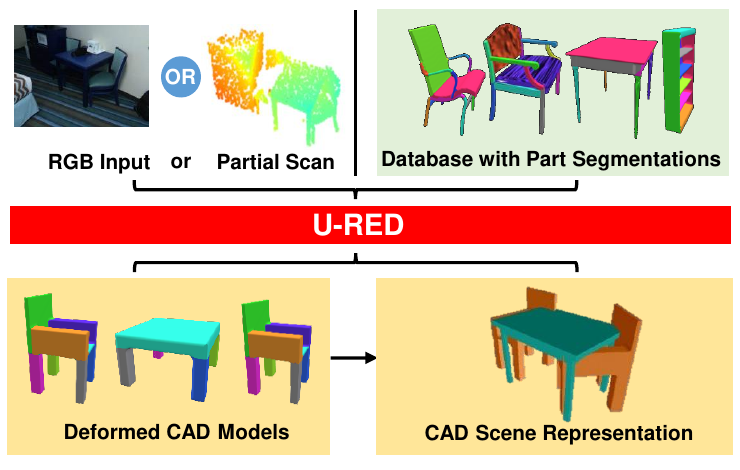}
    \caption{Given a segmented RGB image with estimated depth or a partial noisy object scan, U-RED utilizes an unsupervised joint \textbf{R$\&$D} network to retrieve the most suitable CAD model from the database and deform it to tightly fit the target object. After aligning all deformed shapes to the target scene via predicted poses~\cite{gumeli2022roca}, a compact CAD scene representation is generated.}
    \label{fig:teaser}
%\vspace{-0.5cm}
\end{figure}

3D semantic scene perception~\cite{nie2020total3dunderstanding, zhang2021holistic} involves the decomposition of a scene into its constituent objects, understanding and reconstructing all the detected objects, and putting them in place to formulate a holistic scene representation.
Significant progress has been made recently in attaining the comprehensive analysis of multiple objects' geometry~\cite{sun2021neuralrecon,meshrcnn,niceslam,liu2022planemvs,zhang2022sst},  dynamic reconstruction of both scene and objects~\cite{di2019monocular,di2020unified}, structure-aware scene completion~\cite{cao2022monoscene}, \textit{etc}.
These methods demonstrate promising results in overall reconstruction quality but typically fail in preserving fine-grained geometric structures.
To address this problem, \textbf{Retrieval} and \textbf{Deformation} (\textbf{R$\&$D}) methods~\cite{dahnert2019joint,uy2021joint, nan2012search, uy2020deformation,yifan2020neural,wang20193dn,jiang2020shapeflow,schulz2017retrieval} are proposed.
They leverage a pre-prepared 3D shape database (usually represented as CAD models) as prior knowledge and typically follow a two-stage scheme to generate a clean and compact scene representation.
First, based on manually picked metrics, the most similar shape of the target is selected from the database.
Then, the retrieved shape is scaled, aligned and rotated to match the target.

However, these methods suffer from two challenges, making them vulnerable to noise and partial observations.
First, a partial shape may correspond to multiple full shapes.
 For example, if only a plane is observed, it may be the back or the seat of a chair.
Without additional prior information, the correspondence is totally ambiguous.
Directly applying supervision using a single ground truth may result in erroneous or undesired results.
Therefore, the retrieval network should allow a \textit{one-to-many} (\textbf{OTM}) retrieval.
Second, due to challenging illumination conditions and inherent sensor limitations, noisy observations are common in real-world scenarios.
Thereby, a learned noise-robust metric to measure similarity among shapes is essential for retrieving the most similar source shapes to the observed shape.
% Moreover, geometric consistencies resistant to disturbance of noise should be enforced during joint training of retrieval and deformation.

To handle these challenges, we propose U-RED, a novel \textbf{U}nsupervised joint 3D shape \textbf{RE}trieval and \textbf{D}eformation framework that is capable of effectively handling noisy, partial, and unseen object observations. 
We develop U-RED upon large-scale synthetic data that provides plenty of high-quality CAD models. 
We simulate real-world occlusions, sensor noise, and scan errors to generate partial point clouds of each shape as our network's input for training, then directly apply our method to challenging real-world scenes without fine tuning, since collecting 3D annotations in real scenes is laborious and requires considerable expertise.

To enable \textit{one-to-many} retrieval, we propose to encapsulate possible full shapes of the target partial observation in the surface of a high-dimensional unit sphere.
Specifically, during joint training, a supplementary branch for processing the full shape is incorporated to extract the normalized global feature, which corresponds to a point on the surface of the sphere.
The full-shape feature is concatenated with the target partial-shape feature as an indicator for individual retrieval.
In this manner, the retrieval network learns to interpolate different full shapes on the sphere.
During inference, we sample uniformly on the sphere surface to yield multiple retrievals and collect unique ones as the final results.
Moreover, cross-branch geometric consistencies can be established based on the joint learning scheme to help the partial branch learn structure-aware features and improve robustness against noise.

For similarity metrics, existing methods like~\cite{uy2021joint} directly estimate a single probabilistic score based on Chamfer Distance.
However, this score depends heavily on the training set and is vulnerable to noise due to the unstable nearest neighbor search in Chamfer Distance~\cite{GIANNELLA2021106115}, limiting the generalization ability of these methods, especially in real-world scenes.
We take a step further by designing a novel point-wise residual-guided metric.
For each point inside the target partial observations, we predict a residual vector describing the discrepancy between the coordinates of its own and its nearest neighbor in the source shape.
By aggregating all residual vectors and removing outliers, we calculate an average norm of the remaining vectors as the final metric.
We demonstrate that our residual-guided metric is robust to noise and can be directly applied to real-world scenes while trained only with synthetic data.

Our main contribution are summarized as follows:
\begin{itemize}
\setlength{\itemsep}{0pt}
\setlength{\parsep}{0pt}
\setlength{\parskip}{1pt}
\item 
U-RED, a novel unsupervised approach capable of conducting joint 3D shape \textbf{R$\&$D} for noisy, partially-observed and unseen object observations in the real world, yielding state-of-the-art performance on public synthetic PartNet~\cite{mo2019partnet}, ComplementMe~\cite{sung2017complementme} and real Scan2CAD \cite{avetisyan2019scan2cad} datasets.
\item
A novel \textbf{OTM} module that leverages supplementary full-shape cues to enable \textit{one-to-many} retrieval and enforce geometric consistencies.
\item
A novel \textbf{Residual-Guided Retrieval} technique that is robust to noisy observations in real-world scenes.
\end{itemize}
\section{Related Work}
\textbf{3D Shape Generation and Representation.}
Many deep learning based methods have been proposed to formulate compact representations for 3D shapes in latent space. 
\cite{chen2019learning,mescheder2019occupancy,park2019deepsdf, remelli2020meshsdf,mildenhall2021nerf,xu2022point,jang2021codenerf} try to construct an implicit function by neural networks. 
\cite{achlioptas2018learning,yang2019pointflow,mo2019structurenet,sun2020pointgrow} adopt generative models to produce point clouds with high quality. 
Some prior works \cite{li2017grass,gao2019sdm} also present techniques of factorized representations of 3D shapes, where structural variations are modeled separately within every geometric part.
Another common line of works for 3D shape representation learning \cite{dai2017shape,rao2022patchcomplete,jiang2022masked,yang2021single,xie2019pix2vox} utilize encoder-decoder networks to generate high-dimensional latent codes which contain geometric and semantic information. 
A simple solution of shape retrieval task \cite{li2015joint, uy2021joint,gumeli2022roca} is to directly compare the similarity of source and target shapes in latent space generated by encoder-decoder networks. 
However, such method is extremely sensitive to the quality of the target shapes and hard to handle partial and noisy point clouds. 
%Thus, we design an elaborated retrieval module to effectively utilize latent embeddings and gain robust results.

\textbf{CAD Model Retrieval.}
Retrieving a high-quality CAD model which matches tightly with a real object scan has been an important problem for 3D scene understanding. 
Prior static retrieval works consider the CAD-scan matching task as measuring similarity in the descriptor space \cite{schulz2017retrieval, bosche2008automated} or the latent embedding space as encoded by deep neural networks \cite{li2015joint,dahnert2019joint, gumeli2022roca, avetisyan2019end}. 
In contrast, other approaches \cite{nan2012search} model this task as an explicit classification problem. 
Since database shapes could possess the best fitting details after undergoing a deformation step, \cite{uy2020deformation} proposes to extract deformation-aware embeddings, and \cite{ishimtsev2020cad} designs a novel optimization objective.  
However, these retrieval methods ignore the inherent connection of retrieval and deformation, leading to accumulated error and inferior performance. 
%Thus, we adopt a collaborative learning strategy for \textbf{R\&D} and optimize the two modules simultaneously.

\begin{figure*}[t]
    \centering
    \includegraphics[width=0.97\textwidth]{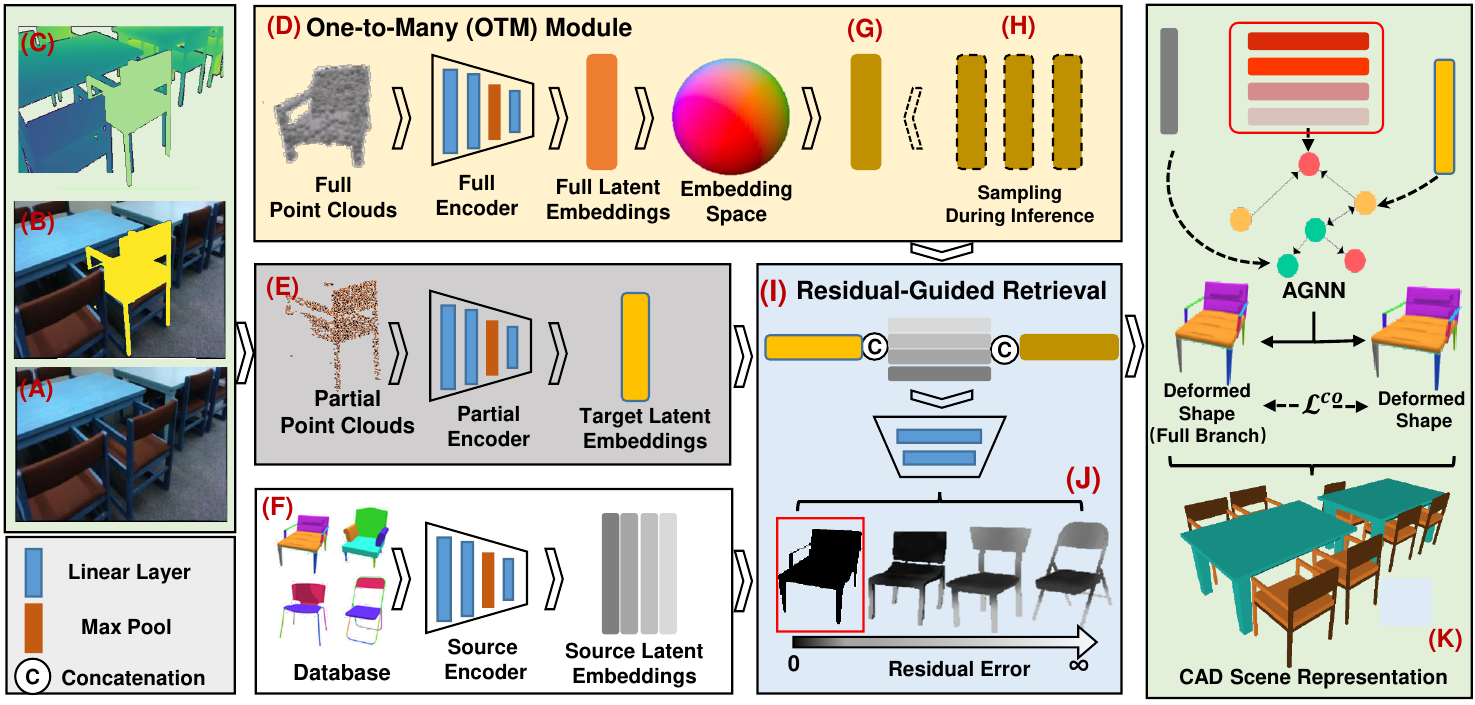}
     \caption{\textbf{Overview of U-RED.}
    Given an RGB image (A) that captures the target scene,
    we leverage an off-the-shelf object detector~\cite{he2017mask} to detect target objects (B) and a depth estimator~\cite{gumeli2022roca} to predict the depth (C). 
    (C) can also be directly attained via scanning.
    We then utilize an arbitrary pose estimator~\cite{gumeli2022roca,di2022gpv,zhang2022rbp,zhang2022ssp} to roughly calibrate the object point cloud.
    Thereby partial point cloud is obtained and fed into our partial encoder (E) to extract the target feature.
    Subsequent retrieval module (I) takes in the target feature, source embeddings from (F) and normalized full shape feature (G), and outputs the residual field $R$ (J) for each source shape.
    Note that in training, (G) is obtained via extracting latent embedding from the supplementary full shape branch, while during inference, (G) represents random samplings on the surface of a unit sphere.
    We choose the source shape with minimum \textit{mean($R$)} or \textit{max($R$)} as the best-fit model.
    Note that each sampling (G) yields a retrieval, we posit the source shape that is selected the most times as the final result.
    Then the retrieved source shape feature, together with the target feature and part features (in red), are optimized in AGNN~\cite{thekumparampil2018attention}.
    The optimized part features are finally passed to an MLP to predict bounding boxes of each part.
    We align all the objects to generate a compact CAD-model-based scene representation.
    }
    \label{fig:pipeline}
\vspace{-0.2cm}
\end{figure*}

\textbf{3D Shape Deformation.}
Traditional methods proposed in the computer graphics community \cite{sorkine2007rigid,huang2008non,ganapathi2018parsing} directly optimize the deformed shapes to fit the input targets. 
However, these approaches commonly struggle to deal with real noisy and partial scans. 
Neural network based techniques instead try to learn deformation priors from a collection of shapes, representing deformation as volumetric warps~\cite{kurenkov2018deformnet, jack2018learning}, cage deformations \cite{yifan2020neural}, vertex-based offsets \cite{wang20193dn} or flows \cite{jiang2020shapeflow}. 
These deformation techniques usually require constraints on grid control points, cage meshes, or number of vertices, which make them less suitable for databases with geometrically heterogeneous parts. 
Moreover, these assumptions are often hard to satisfy in real scans under noisy and heavily occluded settings. 
Recently, \cite{uy2021joint} proposes a novel training strategy with a combined loss to jointly optimize \textbf{R$\&$D} at the same time. 
Although gaining decent performance on a synthetic dataset, its generalization ability to the partially-observed noisy scans in the real world is shown to be limited. 
We propose an unsupervised collaborative training technique and more tightly-coupled design for retrieval and deformation modules, yielding 3D shapes with higher quality and more precise details even when handling partial and noisy point clouds.

\section{Method}

\subsection{Overview}
Given an RGB image $I$ or a scan $\mathcal{S}$ that captures the target scene, our method aims to detect, retrieve and deform all of its objects $O=\{O_1, ...O_N\}$, where $N$ denotes the number of objects, and generate a clean and compact mesh-based scene representation (Fig. \ref{fig:pipeline}).

\noindent \textbf{Feature Extraction.}
We stack three parallel feature encoders to extract $\{\mathcal{F}^{p} \in \mathbb{R}^{M\times L}, \mathcal{G}^{p} \in \mathbb{R}^{L}$\} for the target partial point cloud $\mathcal{T}^p\in \mathbb{R}^{M\times 3}$,  $\{\mathcal{F}^{f} \in \mathbb{R}^{M\times L}, \mathcal{G}^{f}\in \mathbb{R}^{L}\}$ for the corresponding full shape $\mathcal{T}^f\in \mathbb{R}^{M\times 3}$ (Fig.~\ref{fig:pipeline} (D), only in training) and $\{\mathcal{F}^{d} \in \mathbb{R}^{M\times L_d}, \mathcal{G}^{d}\in \mathbb{R}^{L_d}\}$ for sources shapes $\mathcal{O}^c$ in the database.
Here $\mathcal{F}^{*}$ denotes the point-wise feature and $\mathcal{G}^{*}$ encapsulates global information.
$M$ is the number of points and $\{L, L_d\}$ are feature dimensions.
Part features $\{\mathcal{P}^{f}_i \in \mathbb{R}^{L}, i=1,2,...,N_p\}$ of $\mathcal{O}^c$ are computed by mean pooling $\{\mathcal{F}^{d}\}$ with the given part segments.
$N_p$ denotes the number of parts.

\noindent \textbf{Retrieval.} (Sec.~\ref{RGR})
The input is the concatenation of partial features $\{\mathcal{F}^{p}, \mathcal{G}^{p}\}$, source shape features $\mathcal{G}^d$ and the normalized full shape indicator $\mathcal{\hat{G}}^f$ from $\mathcal{G}^f$ in training or a sampling $\mathcal{G}^s$ on the surface of the unit sphere $\Omega$ in inference (Fig.~\ref{fig:pipeline} (G, H)).

\noindent \textbf{Deformation.} (Sec.~\ref{GABD})
Our deformation network consists of an AGNN for part-aware message propagation and an MLP for regressing the bounding box of each part.
AGNN takes $\{\mathcal{P}^{f}, \mathcal{G}^{d}\}$ of the retrieved shape, as well as the target feature $\mathcal{G}^{p}$ as input and outputs updated part features $\mathcal{P'}^{f}$, which are then fed into the MLP to predict per-part bounding boxes.
The input source shape $\mathcal{O}_c$ is finally deformed to $\mathcal{\Tilde{O}}^c$ that tightly matches the target $\mathcal{T}^p$ (Fig.~\ref{fig:pipeline} (K)).

\begin{figure}[t]
    \centering
    \includegraphics[width=0.45\textwidth]{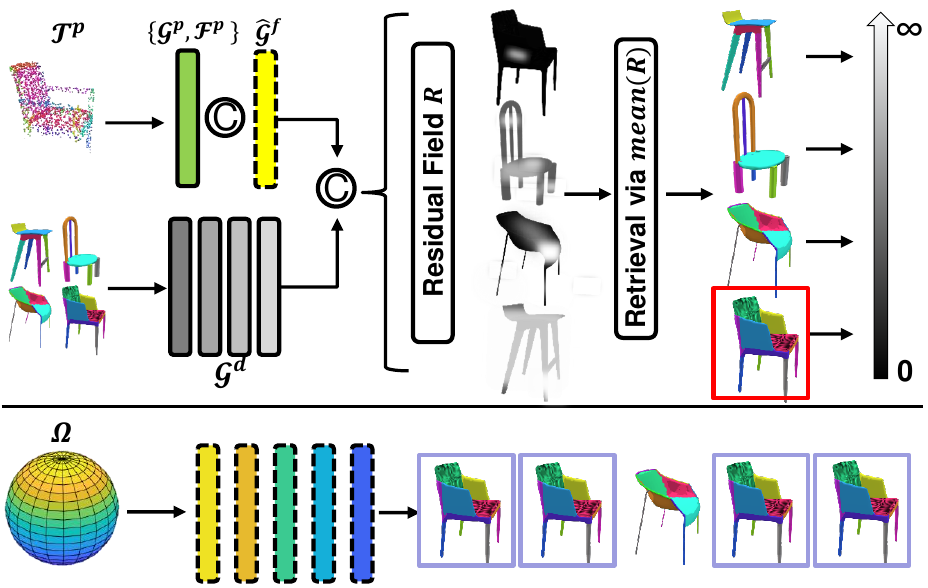}
    \caption{\textbf{Partial point cloud retrieval.} 
    Given target partial features $\{\mathcal{G}^p, \mathcal{F}^p\}$ and candidate source shape features $\mathcal{G}^{d}$, we use the normalized full shape feature $\mathcal{\hat{G}}^{f}$ as the indicator during training, while in inference we randomly sample a feature vector from the unit sphere $\Omega$.
    We estimate point-wise residual field $R$ for each source shape, and select the object with minimum \textit{mean}($R$) (chair in the red square).
    Note that each sampled vector yields a retrieval result, as shown in the bottom block, we choose the object that is selected the most times in sampling as the final retrieval result (chair in the purple square).}
    \label{fig:method}
\end{figure}

\subsection{Retrieval}\label{RGR}
It is essential for the retrieval network to learn a latent metric to measure the similarity between the target $\mathcal{T}^p$ and source shapes $\mathcal{O}_c$.
Compared with previous methods leveraging full shape that encapsulates complete structures of the target , partial shape retrieval puts forward two problems.

The first problem is that one partial shape may correspond to multiple full shapes, yielding multiple feasible retrievals.
Thus the retrieval network must learn a \textit{one-to-many} relationship that enables retrieval of all possible source shapes of the target.
For full shape retrieval, this can be conveniently implemented by simply employing the Chamfer Distance to supervise shape deformation and retrieval~\cite{uy2020deformation, uy2021joint}.
However, for partial point clouds, directly applying the Chamfer Distance may lead to inaccurate nearest neighbor search and thus result in erroneous supervision.
Exemplary, if only one stick is observed, it can originate from the back of a chair, yet, it can also represent a chair leg.
Thereby different interpretations of the observation can lead to very different retrieval results.
A straightforward idea is to harness generative point cloud completion methods~\cite{zhang2021unsupervised, yu2021pointr, huang2020pf} to attain full point clouds.
However, this may significantly increase the inference time and limit real-world applications.

The second problem is that observed point clouds typically contain noticeable noise due to inherent sensor limitations or segmentation errors, which raises a requirement for certain robustness of the retrieval metric.
Existing methods~\cite{gumeli2022roca, uy2021joint, uy2020deformation} first extract global features for the target and source shapes respectively, and then calculate retrieval probability by incorporating certain distance metrics between the features.
Such strategies depend heavily on the training datasets and are vulnerable to real-world noise.

In this module, we introduce two novel techniques to solve the aforementioned problems.

To facilitate the \textit{one-to-many} retrieval, we design a novel \textbf{OTM} module that learns to project the space of all potential full shapes $\mathcal{T}^f$ of the target partial shape $\mathcal{T}^p$ onto the surface $\Omega$ of a high-dimensional unit sphere.
Specifically, during training, a supplementary branch for processing $\mathcal{T}^f$ is incorporated to extract full shape features $\{\mathcal{F}^f, \mathcal{G}^f\}$.
$\mathcal{G}^f$ is normalized to be $\mathcal{\hat{G}}^f$, which corresponds to a point on $\Omega$.
Given the partial shape features $\{\mathcal{F}^p, \mathcal{G}^p\}$, we concatenate them with $\mathcal{\hat{G}}^{f}$ as the input of the retrieval network.
$\mathcal{\hat{G}}^{f}$ serves as an indicator enabling \textit{one-to-one} retrieval.
The network learns to interpolate on $\Omega$ so that each point on $\Omega$ implies a possible full shape.
During inference, we uniformly sample surface points on $\Omega$ as the indicator and each sampling will yield a retrieval result.
We collect the unique retrievals as all feasible results and the source shape that is selected the most times is considered as the best-fit retrieval.

To robustly handle the noisy observations, we design a novel point-wise \textbf{Residual-Guided} similarity metric for \textbf{Retrieval}, as shown in Fig.~\ref{fig:method}.
Our retrieval network predicts the residual field $R=\{R_i \in \mathbb{R}^{3}, i=1,...,M\}$ where for each point $P_i$ in $\mathcal{T}^p$, its corresponding residual vector $R_i$ describes the displacement vector from $P_i$ to its nearest neighbor $O_i$ in the deformed source shape $\mathcal{\Tilde{O}}^c$. 
We adopt the $\mathcal{L}_2$ loss to supervise the training of $R$ with
\begin{equation}
\mathcal{L}^{re} = \frac{1}{M} \sum_{i=1}^{M} \|P_i + R_i - Q_i\|^2,
\label{LRE}
\end{equation}
where $Q_i$ is obtained using nearest neighbor search.
Leveraging $R$, we can derive $mean(R)$ for retrieval.
To be robust to noise, we sort $R$ by norm and remove 10$\%$ points with large residual norms, and then compute the average norm $mean(R)$ of the remaining points as the final metric.

\subsection{Graph Attention Based Deformation}\label{GABD}
Our deformation network consists of an AGNN~\cite{thekumparampil2018attention} for part-aware message aggregation and a regressor to output the bounding box of each part. 
The AGNN takes in 3 types of nodes.
The first is global features $\mathcal{G}^{d}$ from the retrieved source shape $\mathcal{O}_c$.
The second is global features $\mathcal{G}^{p}$ from the target object $\mathcal{T}^p$.
The last is part features $\{\mathcal{P}^{f}_i, i=1,2,...,N\}$ of $\mathcal{O}^c$.
We stack two interleaving self-attention~\cite{vaswani2017attention} and cross-attention modules.
In self attention, different parts exchange information, while in cross attention, global nodes propagate global structure information to guide part nodes.
%Every pair of nodes is connected with a bi-directional edge for message passing.
The overall update process is defined as,
\begin{equation}
\mathcal{F'} = \mathcal{F} + MHA(\mathcal{Q}, \mathcal{K}, \mathcal{V})
\end{equation}
where $MHA$ refers to multi-head attention mechanism~\cite{vaswani2017attention}.
In self-attention modules, $\mathcal{Q},\mathcal{K},\mathcal{V}=\mathcal{F}=\mathcal{P}^f$, while in cross-attention modules, $\mathcal{Q}=\mathcal{F} = \mathcal{P}^f$, $\mathcal{K},\mathcal{V}$ = $\{\mathcal{G}^p, \mathcal{G}^d\}$, the concatenation of global features.

Finally, $\mathcal{P}^{f}$ is fed into the regressor to predict center displacement $\mathcal{C}_d$ and axis-aligned scaling parameters $\{s_w,s_h,s_l\}$.
The final bounding box of each part is recovered as $\mathcal{C}=\mathcal{C}_d + \mathcal{C}_0$, $\{\mathcal{H}, \mathcal{W}, \mathcal{L}\}= \{s_w\mathcal{W}_0, s_h\mathcal{H}_0, s_l\mathcal{L}_0\}$, where $\{\mathcal{C}_0, \mathcal{W}_0, \mathcal{H}_0, \mathcal{L}_0\}$ are initial bounding box center, width, height and length respectively.

\subsection{Joint Training for R\&D}
\label{UCGT}
We utilize the synthetic dataset, PartNet~\cite{mo2019partnet} to generate the training data.
For each available full CAD shape, we simulate real-world partial observations by random cropping and noise addition.
Please refer to the Supplementary Materials for details about data generation.
After training on the simulated data, we directly apply our U-RED on real-world scenes without finetuning.

\iffalse
\textbf{Supplementary Full-Shape Branch.}
As introduced in Sec.~\ref{RGR}, we incorporate an additional branch to process full shapes to enable \textit{one-to-many} retrieval.
Besides this role, the full-shape branch can also be utilized to guide the partial-shape branch to understand the inherent geometric cues of the partial observations (Fig.~\ref{fig:pipeline} (E)).
We use the same network architecture as the partial branch to retrieve and deform geometrically similar shapes from the database for the full shape.
Then geometrically consistency terms can be established between the two branches by means of regularization.
Note that the full-shape branch is only used during training since the ground truth full shape is not available during inference. 
\fi

\textbf{Cross-Branch Consistencies.}
As introduced in Sec.~\ref{RGR} and Fig.~\ref{fig:pipeline} (D), we incorporate an additional branch to process full shapes to enable \textit{one-to-many} retrieval.
Besides this role, the full-shape branch can also be utilized to guide the partial-shape branch to understand the inherent geometric cues of the partial observations (Fig.~\ref{fig:pipeline} (E)).
Thereby our joint learning strategy establishes explicit geometric consistencies between the two branches.
For the deformation network, the retrieved source shape $\mathcal{O}^c$ is deformed to $\mathcal{\Tilde{O}}^c_p$ by the partial-shape branch and $\mathcal{\Tilde{O}}^c_f$ by the full shape branch.
$\mathcal{O}^c_p$ should be consistent with $\mathcal{O}^c_f$.
Meanwhile, given deformed source $\mathcal{\Tilde{O}}^c_p$ from the partial-shape branch and $\mathcal{\Tilde{O}}^c_f$ from the full-shape branch, our retrieval network yields $R_p$ and $R_f$.
By enforcing the consistencies between two branches, the partial-shape branch is guided to produce similar results as the full-shape branch and forced to exploit geometric characteristics of the full shape from the partial input.
For each point $P_i$ in the partial shape, its residual $\{R_{p_i} \in R_p\}$ should be consistent with the residual $\{R_{q_i} \in R_f\}$ of its corresponding point $Q_i$ in the full shape.
We define $R'_f=\{R_{q_i}\}$, and thus our cross-branch consistency loss is defined as,
\begin{equation}
\mathcal{L}^{co} = \mathcal{L}_1^{co} + \mathcal{L}_2^{co} = \|\mathcal{O}^c_p - \mathcal{O}^c_f\|^2 + \|\mathcal{R}_p - \mathcal{R'}_f\|^2
\end{equation}
where we use the Euclidean Distance between point sets as the loss function.

\textbf{Overall Objective.}
Aggregating all losses used in our \textbf{R$\&$D} framework, the final objective for our unsupervised training can be summarized as,
\begin{equation}
\begin{aligned}
\mathcal{L} = \lambda_{0}\mathcal{L}^b + \lambda_{1}\mathcal{L}^{re} + \lambda_{2}\mathcal{L}^{co} 
\end{aligned}
\end{equation}
where $\{\lambda_{0}, \lambda_{1}, \lambda_{2}\}$ are weighting parameters.
$\mathcal{L}^b$ contains several basic losses, including Chamfer Distance loss $\mathcal{L}^{cd}$ and reconstruction loss $\mathcal{L}^r$.
We provide more details in the Supplementary Material.

\section{Experiments}
We mainly conduct experiments on partial input in this section.
For results of full shape input, please refer to the Supplementary Material.
\subsection{Experimental Setup}
\textbf{Dataset Preparation.}
We evaluate U-RED on 3 public datasets. 
In particular, two synthetic datasets PartNet~\cite{mo2019partnet}, ComplementMe~\cite{sung2017complementme} and one real-world dataset Scan2CAD~\cite{avetisyan2019scan2cad}.
PartNet is associated with ShapeNet~\cite{chang2015shapenet} and provides fine-grained part segments for 3 furniture categories: chairs (6531), tables (7939) and cabinets (1278).
ComplementMe contains automatically-segmented shapes of two categories, \textit{i.e.} chairs and tables.
To further demonstrate the generalization ability and practical utility of U-RED, we also adopt a real-world dataset Scan2CAD which is derived from ScanNet~\cite{dai2017scannet}, to directly evaluate our trained model on synthetic data without finetuning.
We follow~\cite{uy2021joint} and divide PartNet and ComplementMe into database, training and testing splits.
Synthetic partial inputs for training are generated by simulating occlusions and sensor errors on original full CAD shapes in PartNet and ComplementMe.
For Scan2CAD, we adopt the test split of ROCA~\cite{gumeli2022roca} and utilize data of three categories: chairs, tables and cabinets.

\textbf{Database Construction.}
We utilize PartNet and ComplementMe to establish the database.
Following~\cite{uy2021joint}, we randomly selected 10$\%$ of the data from the finest level of PartNet hierarchy and additionally select 
200 chairs and 200 tables from ComplementMe. 
Throughout all experiments, we always rely on the same database.

\textbf{Implementation Details.}
We represent all shapes by uniformly sampling $M = 1024$ points.
We use the AdamW optimizer with initial learning rate $1e-3$ and train U-RED for 200 epochs.
For loss weights, we set $\{\lambda_0, \lambda_1, \lambda_2\} = \{3.0, 0.3, 1.0\}$ unless specified.
For all extracted pointwise, global or part features, we set the feature dimension as $L=L_d = 256$.
The supplementary full-shape branch share the hyperparameter setting with the partial branch.
For retrieval implementation, for fair comparison to Top-K soft retrieval adopted by \cite{uy2021joint}, we also take Top-K source candidates with most selected times as described in Sec. \ref{RGR}.
In all experiments, we sample 1000 times on the unit sphere to generate 1000 retrieval results and use top-10 candidates.

\textbf{Evaluation Metrics.}
We report the average Chamfer Distance on the magnitude of $10^{-2}$ of each category respectively and then estimate the instance-average Chamfer Distance on the whole dataset. 
The Chamfer Distance is calculated between the deformed source shapes and the ground truth shapes.
%For more experimental details we refer the reader to the Supplementary Material.

\textbf{Baseline Methods.}
We train and evaluate all baseline methods~\cite{uy2021joint, gumeli2022roca} and U-RED with the same data split, with our genarated partial shapes.
All baseline methods and their variants are trained with the parameters reported in the original papers until convergence. 
We do not use the inner deformation optimization (IDO) step
of Uy~\textit{et al.}~\cite{uy2021joint} for 2 reasons: 
First, IDO significantly increases the training time by almost 10 times. 
Second, IDO works only on the deformation module as a refinement strategy, and can also be incorporated into U-RED. 
Therefore, we focus on comparing the effectiveness of the basic networks and methods, rather than the results after refinement.

\subsection{Real-world Scenes}
\begin{table}[]
\begin{center}
\begin{tabular}{ccccc}
\hline
Method            & Chair          & Table          & Cabinet  & Average\\
\hline
Uy et al. \cite{uy2021joint} & 3.36 & 6.65 & 7.26 & 4.90
 \\
ROCA* \cite{gumeli2022roca} &4.24  &14.97  & 15.92 & 9.15
 \\
ROCA* \cite{gumeli2022roca}+De &6.99  &8.10  & 13.08 & 8.10
 \\
Ours        & \textbf{2.89} & \textbf{3.16} & \textbf{5.95} & \textbf{3.35}\\
\hline
\end{tabular}
\end{center}
%\vspace{-0.3cm}
\caption{Joint \textbf{R$\&$D} results on real Scan2CAD \cite{avetisyan2019scan2cad} dataset. Our U-RED achieves 31.6\% leap forward over competitors of under the instance-average Chamfer Distance.}
\label{tab:roca_result}
\end{table}

\begin{figure}[t]
    \centering
    \includegraphics[width=0.4\textwidth]{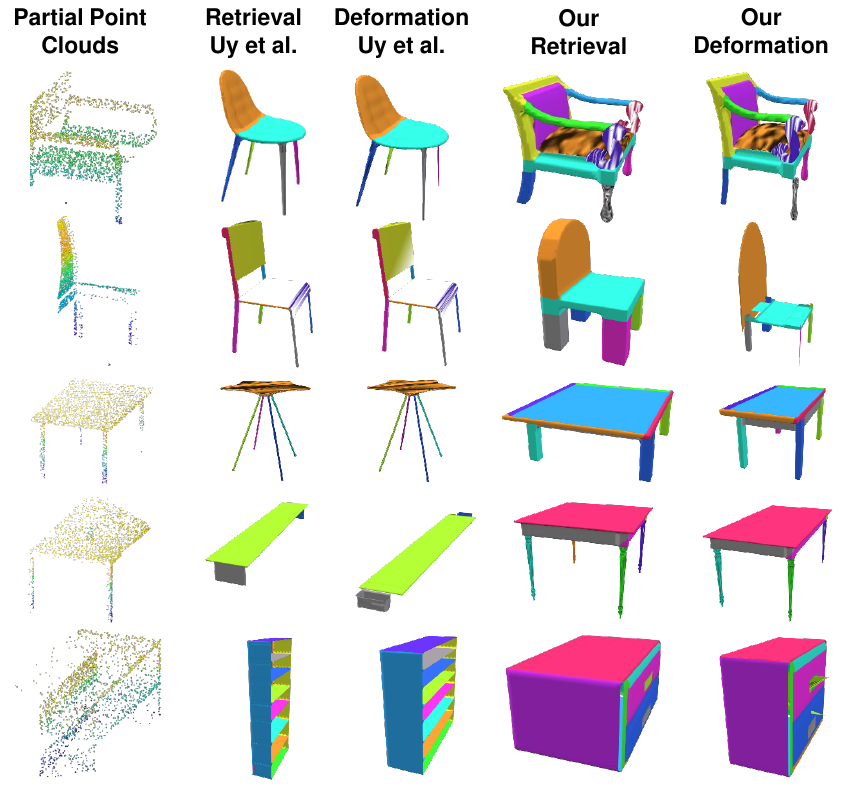}
    \caption{Qualitative comparison on \textbf{R$\&$D} results with Uy \textit{et al.}~\cite{uy2021joint} on real-world Scan2CAD \cite{avetisyan2019scan2cad}.}
    \label{fig:exp_shape_uy}
%\vspace{-0.3cm}
\end{figure}

\begin{figure*}[t]
    \centering
    \includegraphics[width=0.97\textwidth]{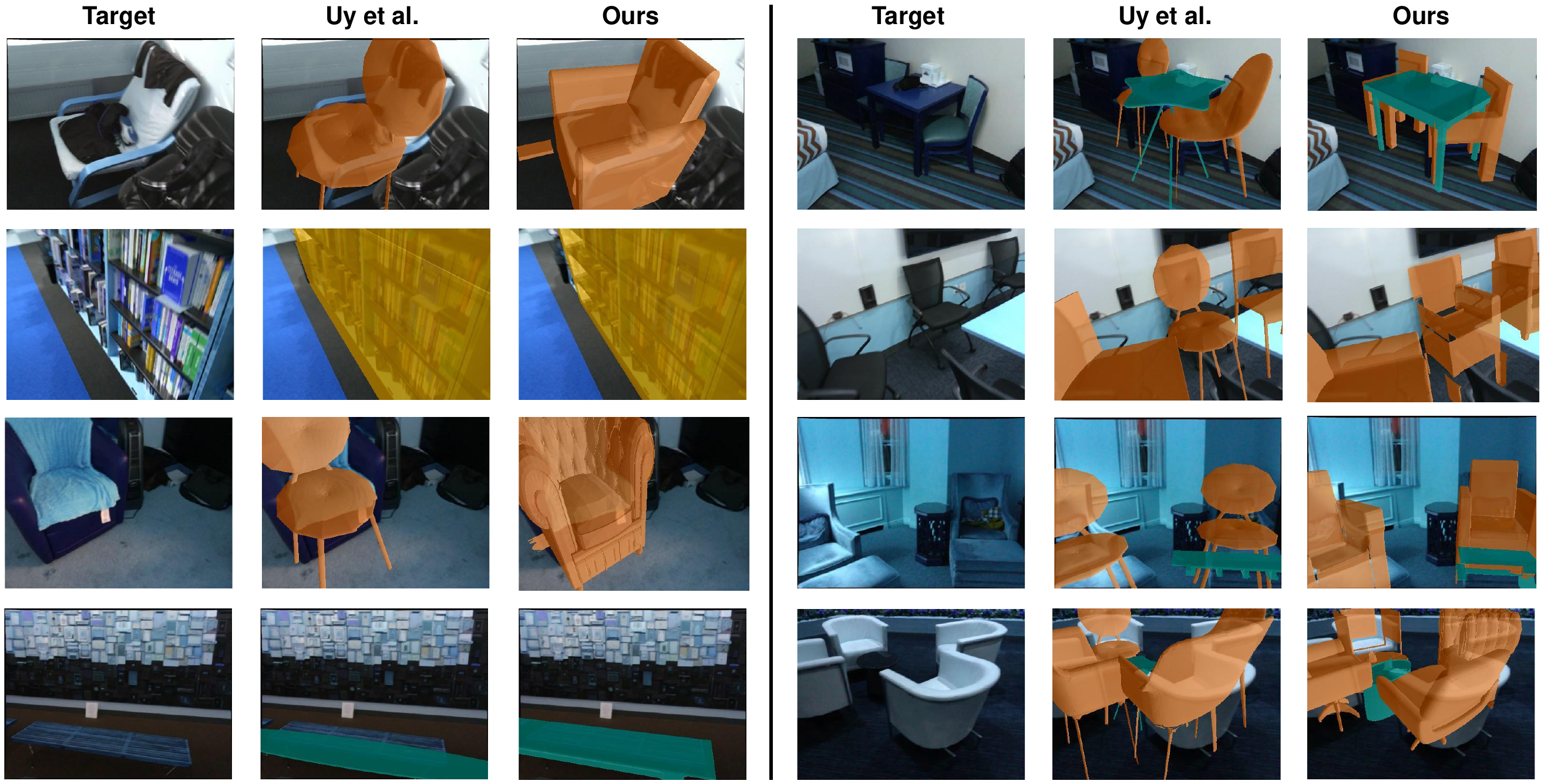}
    \caption{Qualitative results on Scan2CAD \cite{avetisyan2019scan2cad} dataset.
    U-RED consistently outperforms state-of-the-art Uy \textit{et al.}~\cite{uy2021joint}.}
    \label{fig:vis_roca}
\end{figure*}

We first test our proposed approach on the real-world scenes.
We train U-RED on the synthetic PartNet and test it on the real-world Scan2CAD dataset without finetuning or retraining. 
We present the respective results in Tab.~\ref{tab:roca_result}, which demonstrates that U-RED yields the most precise deformed CAD models.
U-RED shows strong generalization ability, which attributes to our elaborated module design and effective unsupervised collaborative learning.

In particular, when comparing with \cite{uy2021joint}, we report superior results on the real-world scenes with decreased Chamfer Distance by 14.0\%, 52.5\% and 18.0\% for three categories respectively. 
Moreover, a qualitative comparison is shown in Fig.~\ref{fig:exp_shape_uy}, and Fig.~\ref{fig:vis_roca}. 
It can be easily seen that our approach yields more accurate retrieval results as well as more precise deformations with respect to the actual real-world objects. Note that for a fair comparison, we train \cite{uy2021joint} with our partial synthetic objects in order to improve its results on partially observed point clouds. 
The noticeable noise in real-world scenes easily deteriorates its performance.

Further, we re-implement ROCA~\cite{gumeli2022roca} by preserving only the retrieval head, notated as ROCA*.
For a fair comparison, we also present results for our improved ROCA*+De, for which we concatenate our deformation head to ROCA*.
We train ROCA* and ROCA*+De on the same synthetic PartNet dataset and test them on Scan2CAD the same as U-RED. 
Compared with ROCA*, the Chamfer Distance yielded by U-RED decreases by 63.3\% on average. 
When compared to ROCA*+De, we still have 58.6\% performance gain (See Tab.~\ref{tab:roca_result}).  
Although the performance of ROCA* improves somewhat when tested on the synthetic dataset, as shown in Tab. \ref{tab:shapenet_result}, it is incapable of conducting proper domain adaptation. 
Despite adding our deformation head in ROCA*+De, the weak retrieval results significantly increase the difficulty of the deformation module, resulting in only a minor improvement, from 9.15 to 8.10.
In contrast, our residual-guided retrieval technique possesses superior noise resistance and enforces effective learning of geometric cues due to the collaborative training procedure.
Thus, it mitigates the negative effects of noisy input and provides robust results for real-world scans. 

In Fig.~\ref{fig:roca_vis_render}, we demonstrate the qualitative comparisons of Ours \textit{vs} Uy \textit{et al.} on Scan2CAD with only RGB input. 
Specifically, given an RGB image as input, we detect the target objects with Mask-RCNN~\cite{he2017mask} and predict depth with ROCA~\cite{gumeli2022roca}.
The recovered point cloud is then transformed via the estimated pose from ROCA and used as input to the methods.

\subsection{Synthetic Scenes}
\begin{table}[]
\begin{center}
\begin{tabular}{ccccc}
\hline
\multicolumn{5}{c}{PartNet \cite{mo2019partnet}}\\
\hline
Method            & Chair          & Table         & Cabinet &Average\\
Uy et al. \cite{uy2021joint} & 2.02 & 2.32 & 2.63 &2.22
 \\
ROCA* \cite{gumeli2022roca} &2.50  &2.72  & 3.86 &2.72
\\
ROCA* \cite{gumeli2022roca}+De &3.80  &3.87  &2.82 &3.76
 \\
Ours        & \textbf{0.95} & \textbf{1.33} & \textbf{1.30} & \textbf{1.17}\\
\hline
\multicolumn{5}{c}{ComplementMe \cite{sung2017complementme}}\\
\hline
Method            & Chair          & Table         & - &Average\\
Uy et al. \cite{uy2021joint} & 2.08 & 2.66 & - &2.40
\\
Ours        & \textbf{1.68} & \textbf{2.26} & - &\textbf{2.00} \\
\hline
\end{tabular}
\end{center}
%\vspace{-0.3cm}
\caption{Joint \textbf{R$\&$D} results on synthetic datasets. Our U-RED outperforms all competitors in synthetic scenes with both manually and automatically segmentation.}
\label{tab:shapenet_result}
%\vspace{-0.3cm}
\end{table}

\begin{figure}[t]
    \centering
    \includegraphics[width=0.45\textwidth]{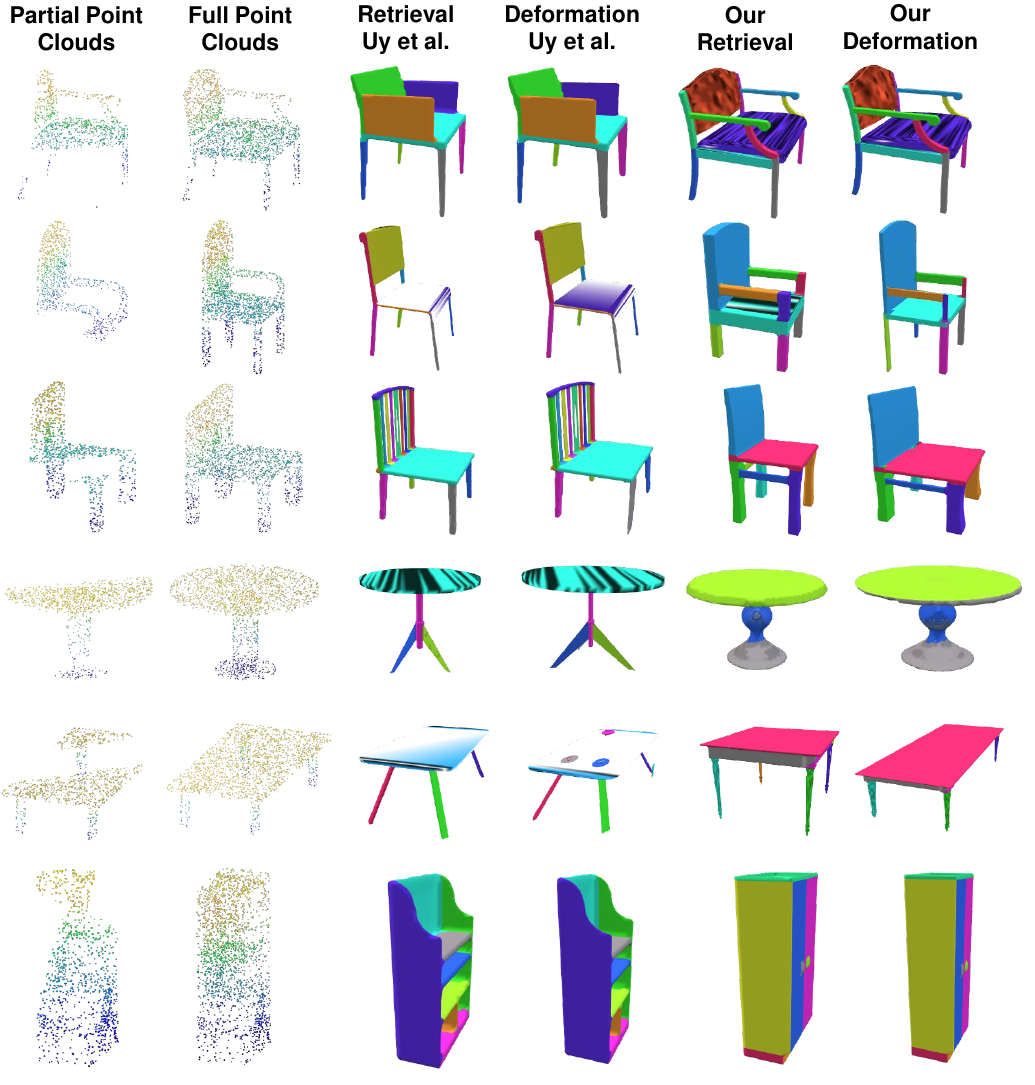}
    \caption{Visualization on PartNet \cite{mo2019partnet}. Qualitative comparison with Uy \textit{et al.} \cite{uy2021joint} demonstrates that our U-RED performs more robustly, gains more accurate retrieval and more precise deformation results when facing partially observed point clouds.}
    \label{fig:exp_shape}
%\vspace{-0.3cm}
\end{figure}

The objective of U-RED is to enhance the generalization ability with unsupervised training procedures to handle noisy and partial input in real-world scenes. 
However, we still conduct several experiments on synthetic datasets for a fair comparison with existing methods \cite{uy2021joint, gumeli2022roca}. 
The experiment results (See Tab.~\ref{tab:shapenet_result}) show that our U-RED also outperforms the state-of-the-art competitors by a large margin when dealing with ambiguous and partially observed point clouds in synthetic scenes.

\begin{figure}[t]
    \centering
    \includegraphics[width=0.45\textwidth]{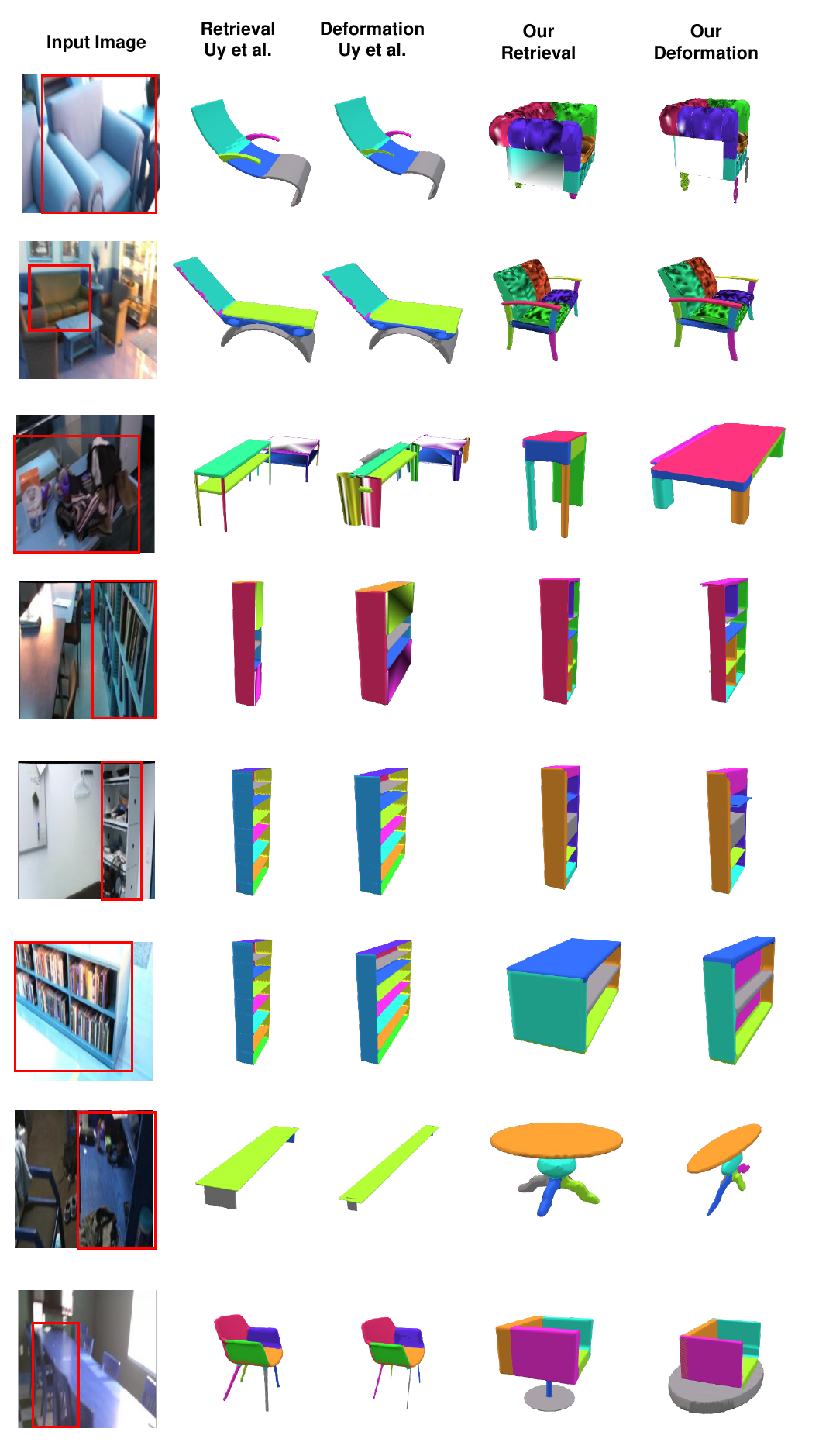}
    \caption{Qualitative results on Scan2CAD \cite{avetisyan2019scan2cad} dataset with only RGB input.
    We leverage an off-the-shelf object detector~\cite{he2017mask} to detect target objects and a depth estimator~\cite{gumeli2022roca} to predict the depth.
    Then we use our U-RED for 3D shape retrieval and deformation. 
    U-RED consistently outperforms state-of-the-art Uy \textit{et al.}~\cite{uy2021joint} considering visual effects.}
    \label{fig:roca_est}
\end{figure}

\begin{figure}[t]
    \centering
    \includegraphics[width=0.45\textwidth]{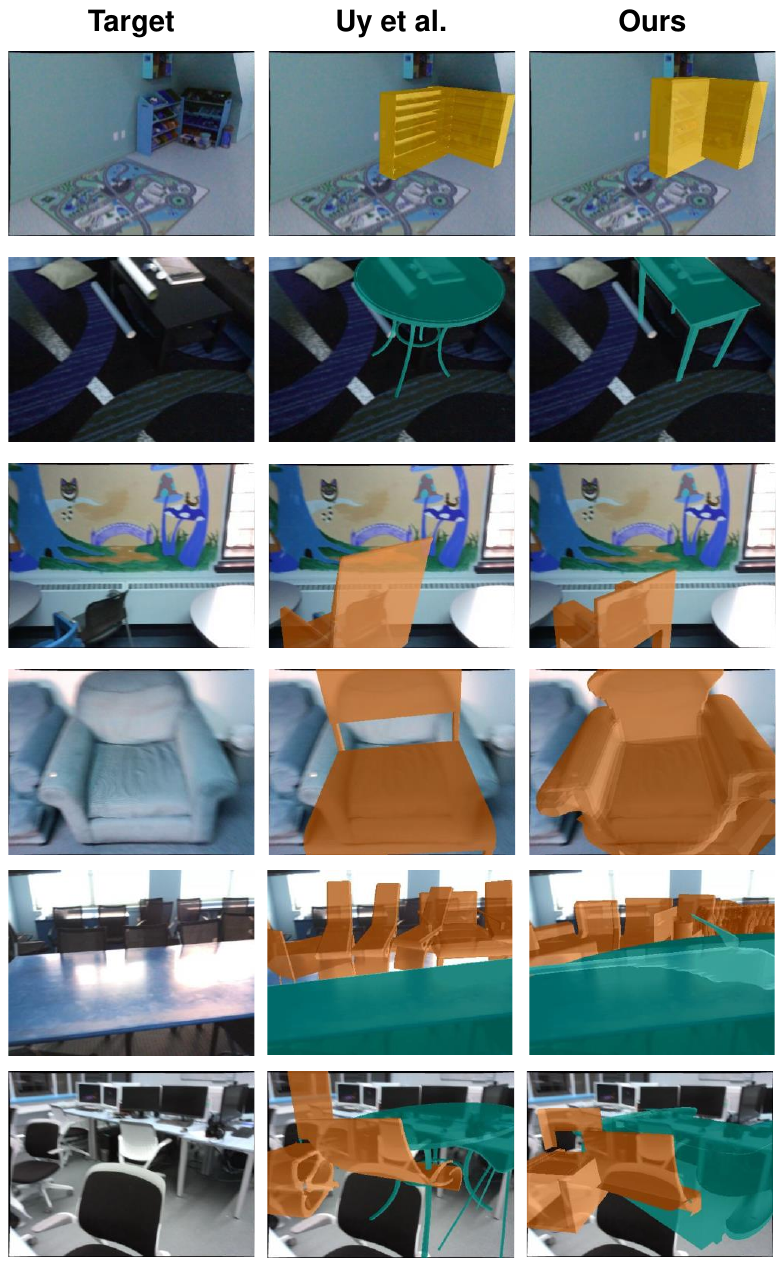}
    \caption{Qualitative results on Scan2CAD \cite{avetisyan2019scan2cad} dataset with only RGB input.
    U-RED consistently outperforms state-of-the-art Uy \textit{et al.}~\cite{uy2021joint}.}
    \label{fig:roca_vis_render}
\vspace{-0.2cm}
\end{figure}

On PartNet, U-RED surpasses \cite{uy2021joint} by 47.3\% on average of the three categories, and exceeds ROCA* and  ROCA*+De by an even larger margin.   
On ComplementMe, U-RED still surpasses \cite{uy2021joint} by 19.2\% and 15.0\% for chair and table categories respectively. 
Fig. \ref{fig:exp_shape} exhibits the qualitative visualization comparison between our U-RED and \cite{uy2021joint}. 
U-RED demonstrates more accurate retrieval results (the chair in the $1^{st}$ row) and stronger deformation capability (the table in the $5^{th}$ row). 
Fig.~\ref{fig:roca_est} demonstrates the results with only RGB input, with the same setting as in Fig.~\ref{fig:roca_vis_render}.

\subsection{Ablation Studies}

\begin{table}[]
\begin{center}
\begin{tabular}{ccccc}
\hline
\multicolumn{5}{c}{Scan2CAD \cite{avetisyan2019scan2cad}}\\
\hline
Method            & Chair          & Table           & Cabinet &Average\\
Uy et al. \cite{uy2021joint} & 2.83 & 2.47 & \textbf{4.56} &2.92
 \\
Ours        & \textbf{2.05} & \textbf{1.43} & 5.27 &\textbf{2.24}\\
\hline
\multicolumn{5}{c}{PartNet \cite{mo2019partnet}}\\
\hline
Method            & Chair          & Table           & Cabinet &Average\\
Uy et al. \cite{uy2021joint} & 1.43 & 1.48 & 1.79 &1.48
 \\
Ours        & \textbf{0.70} & \textbf{0.69} & \textbf{0.74} &\textbf{0.70} \\
\hline
\end{tabular}
\end{center}
%\vspace{-0.3cm}
\caption{Deformation results with oracle retrieval. Results on two datasets show our U-RED gains stronger deformation ability due to our graph attention based deformation module.}
\label{tab:deform_result}
%\vspace{-0.3cm}
\end{table}

\begin{table}[]
\begin{center}
\begin{tabular}{ccccc}
\hline
\multicolumn{5}{c}{Scan2CAD \cite{avetisyan2019scan2cad}}\\
\hline
Method            & Chair          & Table           & Cabinet &Average\\
Pro-Re \cite{uy2021joint}  &7.39  &6.00  &15.24  &5.95\\
Ours w/o \textbf{OTM}   & 3.19 & 6.62 & 7.02 &4.71\\
Ours        &\textbf{2.89} & \textbf{3.16} & \textbf{5.95} &\textbf{3.35}\\
\hline
\multicolumn{5}{c}{PartNet \cite{mo2019partnet}}\\
\hline
Method            & Chair          & Table           & Cabinet &Average\\
Pro-Re \cite{uy2021joint}  &2.84  &5.17  & 4.66 &4.15
 \\
Ours w/o \textbf{OTM}  & 1.17    & 2.16   & 1.85  & 1.76 \\ 
Ours        &\textbf{0.95}&  \textbf{1.33}& \textbf{1.30} &\textbf{1.17}\\
\hline
\end{tabular}
\end{center}
%\vspace{-0.4cm}
\caption{Retrieval ablations.
Ours w/o \textbf{OTM} refers to our method without the \textbf{\textbf{OTM}} module.
Results illustrate that our residual-guided retrieval gains more precise retrieval for the targets than the probabilistic retrieval adopted by \cite{uy2021joint}.}
\label{tab:retrieval_ablation_result}
%\vspace{-0.3cm}
\end{table}

\begin{table}[]
\begin{center}
\begin{tabular}{ccccc}
\hline
\multicolumn{5}{c}{Scan2CAD \cite{avetisyan2019scan2cad}}\\
\hline
Method            & Chair          & Table           & Cabinet &Average\\
w/o $\mathcal{L}^{co}$  &3.57  &5.56  & 6.54 &4.58
 \\
w/o $\mathcal{L}^{r}$  &4.47  &5.03  &9.11 &5.22\\
w/o $\mathcal{L}^{re}$  &2.95  &5.32  &19.6 &5.75\\
Ours        &\textbf{2.89} & \textbf{3.16} & \textbf{5.95} &\textbf{3.35}\\
\hline
\multicolumn{5}{c}{PartNet \cite{mo2019partnet}}\\
\hline
Method            & Chair          & Table           & Cabinet &Average\\
w/o $\mathcal{L}^{co}$  &1.01&  1.41&  2.25 &1.31
 \\
w/o $\mathcal{L}^{r}$  &1.37&  1.63&  1.51 &1.51\\
w/o $\mathcal{L}^{re}$  &2.02&  1.77&  1.54 &1.86\\
Ours        &\textbf{0.95}&  \textbf{1.33}& \textbf{1.30} &\textbf{1.17}\\
\hline
\end{tabular}
\end{center}
%\vspace{-0.4cm}
\caption{Unsupervised training ablations. The results prove that each single loss term in our unsupervised training procedure contributes to better performance.}
\label{tab:training_ablation_result}
%\vspace{-0.3cm}
\end{table}

\begin{table}[]
\begin{center}
\begin{tabular}{ccccc}
\hline
Occlusion          & Chair          & Table           & Cabinet &Average\\
\hline
0\%  &0.77  &1.33  &1.18  &1.08\\
25\%  &0.86  &1.33  &1.23  &1.12\\
50\%  &0.95  &1.33 &1.30 &1.17\\
75\%  &1.09  &2.21 &1.65 &1.69\\
\hline
\end{tabular}
\end{center}
%\vspace{-0.4cm}
\caption{Occlusion ablations on PartNet \cite{mo2019partnet}. Results illustrate that our U-RED performs stably even under heavy occlusion.}
\label{tab:partial_ablation_result}
\vspace{-0.3cm}
\end{table}

We conduct several ablation studies on deformation ability (Tab.~\ref{tab:deform_result}), retrieval ability (Tab.~\ref{tab:retrieval_ablation_result}), the effectiveness of unsupervised joint training techniques (Tab.~\ref{tab:training_ablation_result}) and robustness to input occlusion proportion (Tab.~\ref{tab:partial_ablation_result}). 
These ablation studies demonstrate the effectiveness of U-RED in handling noisy partial points in both real-world and synthetic scenes.

\textbf{Deformation Ability.} 
In this experiment, we research on the deformation ability alone using oracle retrieval, i.e. the deformation network deforms every source shape in the database to match a target shape and reports the minimum Chamfer Distances between the target shape and all deformed shapes. 
We compare the deformation ability of our graph attention-based deformation module with \cite{uy2021joint}.
Tab.~\ref{tab:deform_result} shows that when only considering deformation ability, 
U-RED surpasses \cite{uy2021joint} by 23.3\% on average on real-world Scan2CAD and 52.7\% on synthetic PartNet.

\textbf{Retrieval Ability.}
We aim to verify the effectiveness of our proposed \textit{one-to-many} retrieval and residual-guided retrieval metric. 
As in Tab.~\ref{tab:retrieval_ablation_result}, Ours w/o \textbf{OTM} refers to persevering the residual-guided metric but turn off the \textbf{\textbf{OTM}} module.
Thereby the normalized full shape feature $\mathcal{\hat{G}}^f$ is not fed into the retrieval network.
Moreover, we need a baseline that doesn't leverage both the two techniques.
For this purpose,  we implemented a network with the probabilistic soft retrieval strategy from \cite{uy2021joint} and leverages other modules the same as U-RED, notated as Probabilistic Retrieval (Pro-Re). 
Tab.~\ref{tab:retrieval_ablation_result} verifies the effectiveness of our proposed residual-guided retrieval module. 
Compare Pro-Re and Ours w/o \textbf{OTM}, it can be easily concluded that the residual-guided metric improves the retrieval accuracy remarkably by 20.8\% on Scan2CAD and 56.8\% on PartNet.
When comparing ours w/o \textbf{OTM} and Ours, it's clear that the \textbf{OTM} retrieval strategy contributes 28.9\% and 33.5\% improvements on Scan2CAD and PartNet respectively.
We provide several qualitative retrieval results to further demonstrate the effectiveness of \textbf{OTM} in the Supplementary Material. 

\textbf{Loss terms.} 
We ablate the effectiveness of each proposed loss term on both real and synthetic scenes and summarize the experiment results in Tab.~\ref{tab:training_ablation_result}.
As illustrated, we proved that each single loss term contributes to better performance.
And finally, the network achieve the best performance by applying of all the loss terms.

\textbf{Input Occlusion Proportion.} 
Tab. \ref{tab:partial_ablation_result} examines the robustness of our method when tackling input partially-observed point clouds with different occlusion proportions.
The occlusion proportions are controlled by our simulation technique adopted to generate the partial point clouds input.
Our U-RED performs stably even with an occlusion ratio of 50\%. 
We observe an accuracy drop for the first time when the occlusion ratio increase to 75\%. 
The result verifies the robustness of our approach.

We also provide experiments analyzing the effects of noise level, and different retrieval metrics: min \textit{mean(R) vs.} min \textit{max(R)} in the Supplementary Material.

\section{Conclusion}
In this paper, we present U-RED, a novel unsupervised framework that takes a partial object observation as input, aiming to retrieve the most geometrically similar shape from a pre-established database and deform the retrieved shape to match the target.
To solve the one-to-many retrieval problem and enhance the robustness to handle noisy partial points, we design a collaborative unsupervised learning technique with the aid of a supplementary full-shape branch, enforcing geometric consistencies.
Further, we propose a residual-guided retrieval module that is robust to noisy observations and a graph attention based deformation module for gaining more precise deformed shapes.
The training technique and all proposed modules are demonstrated to be effective through our exhaustive experiments in both real-world and synthetic scenes.
In the future, we plan to apply our retrieval technique in a wide range of real-world applications, \textit{e.g.} model-based robotic grasping.

\textbf{Acknowledgements.}
This work was partially funded by the German Federal Ministry for Economics and Climate Action (BMWK) under GA 13IK010F (TWIN4TRUCKS) and the EU Horizon Europe Framework Program under GA 101058236 (HumanTech).
This work was also supported by the National Key R\&D Program of China under Grant 2018AAA0102801, National Natural Science Foundation of China under Grant 61827804. 

%%%%%%%%% REFERENCES
{\small
\bibliographystyle{ieee_fullname}
\bibliography{egbib}
}

\end{document}